\documentclass[conference]{IEEEtran}
\IEEEoverridecommandlockouts
\usepackage{cite}
\usepackage{amsmath,amssymb,amsfonts}
\usepackage{algorithmic}
\usepackage{graphicx}
\usepackage{textcomp}
\usepackage{xcolor}
\usepackage{paralist}

\usepackage{hyperref}

\usepackage{booktabs} 

\def\BibTeX{{\rm B\kern-.05em{\sc i\kern-.025em b}\kern-.08em
    T\kern-.1667em\lower.7ex\hbox{E}\kern-.125emX}}
\begin{document}

\title{
Fast Fourier Color Constancy and Grayness Index for ISPA Illumination Estimation Challenge
}

\author{
\IEEEauthorblockN{Yanlin Qian}
\IEEEauthorblockA{
\textit{Tampere University}
\\
yanlin.qian@tuni.fi
}
\and
\IEEEauthorblockN{Ke Chen, Huanglin Yu}
\IEEEauthorblockA{
\textit{South China University of Technology}
}
}

\maketitle

\begin{abstract}
We briefly introduce two submissions to the Illumination Estimation Challenge, in the Int'l Workshop on Color Vision, affiliated to the 11th Int'l Symposium on Image and Signal Processing and Analysis. The fourier-transform-based submission is ranked 3rd, and the statistical Gray-pixel-based one ranked 6th. 
\end{abstract}


\section{Introduction}
Color constancy refers to the property of camera capturing the intrinsic color of objects regardless of the casting scene illumination.  Researchers of color constancy are generally doing two things: First, measuring the illumination estimation as accurate as possible; Second, color correction to realize auto white balance. This work is all about the Step 1 as it is the most demanding part. 

Many advanced deep learning methods (\textit{e.g.} \cite{hu2017cvpr, shi2016eccv, yanlin2017iccv, quasiunsupervisedcc2019}) have been proposed, and they achieved nearly-saturating results over two main-stream color constancy datasets -- Gehler-Shi dataset~\cite{shi2010re} and NUS 8-camera dataset~\cite{cheng14}.  Along with the illumination estimation challenge, a new benchmark, Cube+ Dataset~\cite{banic2017unsupervised} is advertised and applied with more hardness. It is meaningful for us to test some existing methods and see a reboot in experimental results. 

Different to Gehler-Shi and NUS 8-camera dataset, the Cube+ contains more images (up to 1707) with diverse ground truth annotations. Please refer to \cite{banic2017unsupervised} for more official description of this dataset. One noticeable point of the Cube+ is that, only a single camera (Canon EOS 550D) is used for collecting the training and testing data. This can be seemed as an advantage and a disadvantage, simultaneously. On one hand, we now have chance to test any methods independent of varying photometric properties of several sensors. On the other hand, we are blind to the varying camera sensibility. This makes Cube+ ``easier'' for CNN-based methods. 

Given that CNN is a functional tool since its invention and many other teams will use it for the challenge (we guess), we test some different approaches. The first submission (ranked 3rd in the leaderboard) is based on Barron's Fast Fourier Color Constancy (FFCC,~\cite{barron2017fourier}) since its source is publicly available and it is practical in products (\textit{e.g.} Google Photo App). For how FFCC works and its insights, please refer to \cite{barron2017fourier}. 

The other submission is ranked 6th which is out of top three, but we still want to introduce it. It is based on Grayness Index (GI) \cite{yanlin2019arxiv}, which is an advanced version of gray pixel methods \cite{yang2015efficient,yanlin2019vissap}. A comparison of gray pixel methods is given by \cite{yanlin2019icip}. It is \textbf{learning-free}, thus it does not rely on training on the given 1707 images and the corresponding groundtruth. We choose it as it is simple (few lines of Matlab, no learning) and we are curious to see its limitness in a more challenging dataset like Cube+. Please check \cite{yanlin2019arxiv} if you want to find how gray (achromatic) pixels can be identified.  

\section{Methodology}
In this section we describe the principles of the two approaches. Assuming Lambertian model, narrow sensor response and uniform global illumination , the RGB value of a pixel can be expressed as:
\begin{align}
\label{eq:onelight_formation}
I^{c}(p) = W^c(p) \circ L^c(p)
\end{align}
which shows the color value of the channel $c$ at the location $p$ in image $I$ is the product of the surface albedo $W^c(p)$ and the illumination color $L^c(p)$. In Barron's \cite{barron2015convolutional}, the RGB value of a pixel $I(p)$ is transformed into the log-chroma measures:

\begin{align}
\label{eq:logchroma}
u(p)=log(I^g(p)/I^r(p)), v(p)=log(I^g(p)/I^r(b)).
\end{align}

By framing the task of color constancy in Equation~\ref{eq:logchroma}, the global illumination $L$ can be treated as an additive constraint in log-chroma space. Now that we have a 2D spatial localization task, where some window-wise classifiers can be trained using image-groundtruth pairs and then gives maximum activation when it detects illumination on the UV histogram.  Barron proposed convolutional filter on UV histogram in \cite{barron2015convolutional}, and further extended it in FFCC \cite{barron2017fourier} by performing element-wise multiplication in the Fourier space, which allows ``warpped'' input images and faster inference speed. 

Grayness Index \cite{yanlin2019arxiv} addresses color constancy from a different angle. Apply \textit{log} and laplacian-of-gaussian filter $\delta$ on Equation.~\ref{eq:onelight_formation}, we get: 
\begin{align}
\label{eq:gp_formation}
\delta\log I^{c}(p)&=\delta\log W^c(p)+\delta\log L^c(p)
\end{align}
Assuming the illumination is constant over small local neighborhood (the same color and direction), Equation~\ref{eq:gp_formation} simplifies to:
\begin{align}
\label{eq:gp_formation2}
\delta\log I^{c}(p)&=\delta\log W^c(p),
\end{align}
which is the core of gray pixel. $\delta\log I^{r}(p)=\delta\log I^{g}(p)=\delta\log I^{b}(p)$ indicates a perfect gray pixel at at the location~$p$. Based on Equation~\ref{eq:gp_formation2}, Grayness Index detects nearly gray pixels accurately, which reflects illumination.  

Please refer to the original papers \cite{barron2017fourier, yanlin2019arxiv} for more theoretical details and the implementation.  

\section{How we do differently}

\noindent\textbf{FFCC-based method} FFCC contains multiple variants (model A to Q, depending on how many channels are used, thumb or full-size input, photo exif or deep feature).  As the challenge page provides exif information, we adopt model P of FFCC, which is the best branch proved by results on Gehler-Shi Dataset. For deep feature required by Model P, we extract the neural activation from the layer ``fc7'' of 16-layer VGG network, which is pretrained on Place365 \cite{zhou2016places}. Now that we have exif and deep feature for each image of Cube+ dataset, we tuned the hyper parameters of FFCC and trained our model, same as \cite{barron2017fourier}.

On the last submission day we realized we made a mistake. The released testing data do not contain any exif, which we believe our trained model will definitely fail on this testing data. Under limited time, we computed the mean exif matrix of all training images and used it for each testing image.  The trained model P can work, but not to a satisfying degree. We submitted the results given by this model P. 

\noindent\textbf{GI-based method} We did not change anything -- we downloaded the code from the github page\footnote{https://github.com/yanlinqian/Grayness-Index} and used it on testing images. 

\section{Results}

\begin{table}
\scriptsize
\begin{center}
\caption{Illumination Estimation Leaderboard. The notation $\Uparrow(n)$ refers to the method being ranked $n$th. The numbers on Cube+ training data are obtained using 3-fold cross-validation for FFCC and are not required for the submission. We list them for better comprehensive comparison.} 
\label{tab:leaderboard}
  \begin{tabular}{l ccc}
    \toprule 
  &  \multicolumn{3}{c}{Undisclosed Testing Data }  \\
 Method & Median & Mean & Trimean  \\
 \midrule 
$\Uparrow(1)~$Color~Cerberus~\textit{et.al.}\cite{colorcerberus} & 1.51 & 2.65 & 1.64  \\
$\Uparrow(2)~$FFCC Model J (barron)& 1.59 & 2.49 & 1.73 \\
$\Uparrow(3)~$FFCC Model P (our) & 1.64 & 2.93 & 1.77 \\
$\Uparrow(6)~$GI (our) & 2.10 & 6.87 & 2.50 \\
\bottomrule
  \end{tabular}
  \bigbreak
  \begin{tabular}{l ccc}
    \toprule 
  &  \multicolumn{3}{c}{ Cube+ Training Data}  \\
 Method & Median & Mean & Trimean  \\
 \midrule 
FFCC Model P (our) & 0.84 & 1.64 & 1.05 \\
GI (our) & 1.21 & 2.07 & 1.46 \\
\bottomrule
  \end{tabular}
  
\end{center}\vspace{-3mm}
\end{table}

\begin{table}
\scriptsize
\begin{center}
\caption{Results on Gehler-Shi Dataset} 
\label{tab:gehlershi}
  \begin{tabular}{l ccc}
    \toprule 
  &  \multicolumn{3}{c}{3-fold cross validation}  \\
 Method & Median & Mean & Trimean  \\
 \midrule 
FFCC Model P & 0.96 & 1.78 & 1.14 \\
GI  & 1.87 & 3.07 & 2.16 \\
\bottomrule
  \end{tabular}
\end{center}\vspace{-3mm}
\end{table}

\begin{table}[t]
\scriptsize
\begin{center}
\caption{Results on NUS 8-camera Dataset} 
\label{tab:nus}
  \begin{tabular}{l ccc}
    \toprule 
  &  \multicolumn{3}{c}{3-fold cross validation}  \\
 Method & Median & Mean & Trimean  \\
 \midrule 
FFCC Model P & 1.31 & 1.99 & 1.43 \\
GI  & 1.97 & 2.91 & 2.13 \\
\bottomrule
  \end{tabular}
\end{center}\vspace{-3mm}
\end{table}

Table~\ref{tab:leaderboard} gives a simple leaderboard. Our FFCC-based Model Q suffers from the lack of exif for testing data, ranked after the FFCC Model J, which does not rely on exif.  This validates that using a biased exif leads to a worse case. The 3-fold cross-validation of FFCC Model P on Cube+ training data shows again the importance of exif information.

\begin{figure}[t]
\begin{center}
\vspace{-0.75mm}
\begin{tabular}{c cc cc}

\hspace{-15pt}
\vspace{-0.75mm}
\raisebox{2\height}{\rotatebox{0}{{  }}}
&{{\includegraphics[width=2cm,height=1.4cm]{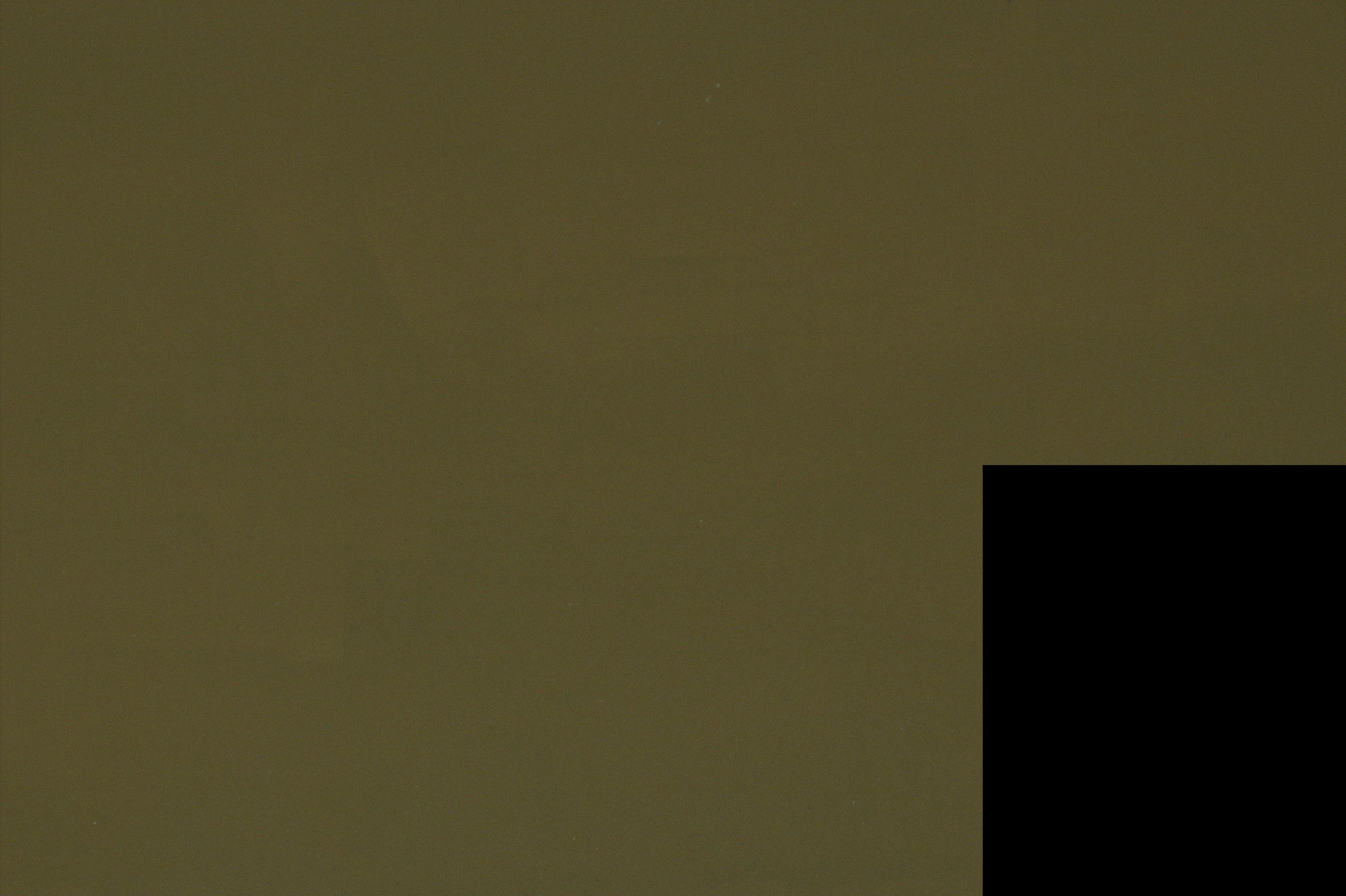}}}
& {{\includegraphics[width=2cm,height=1.4cm]{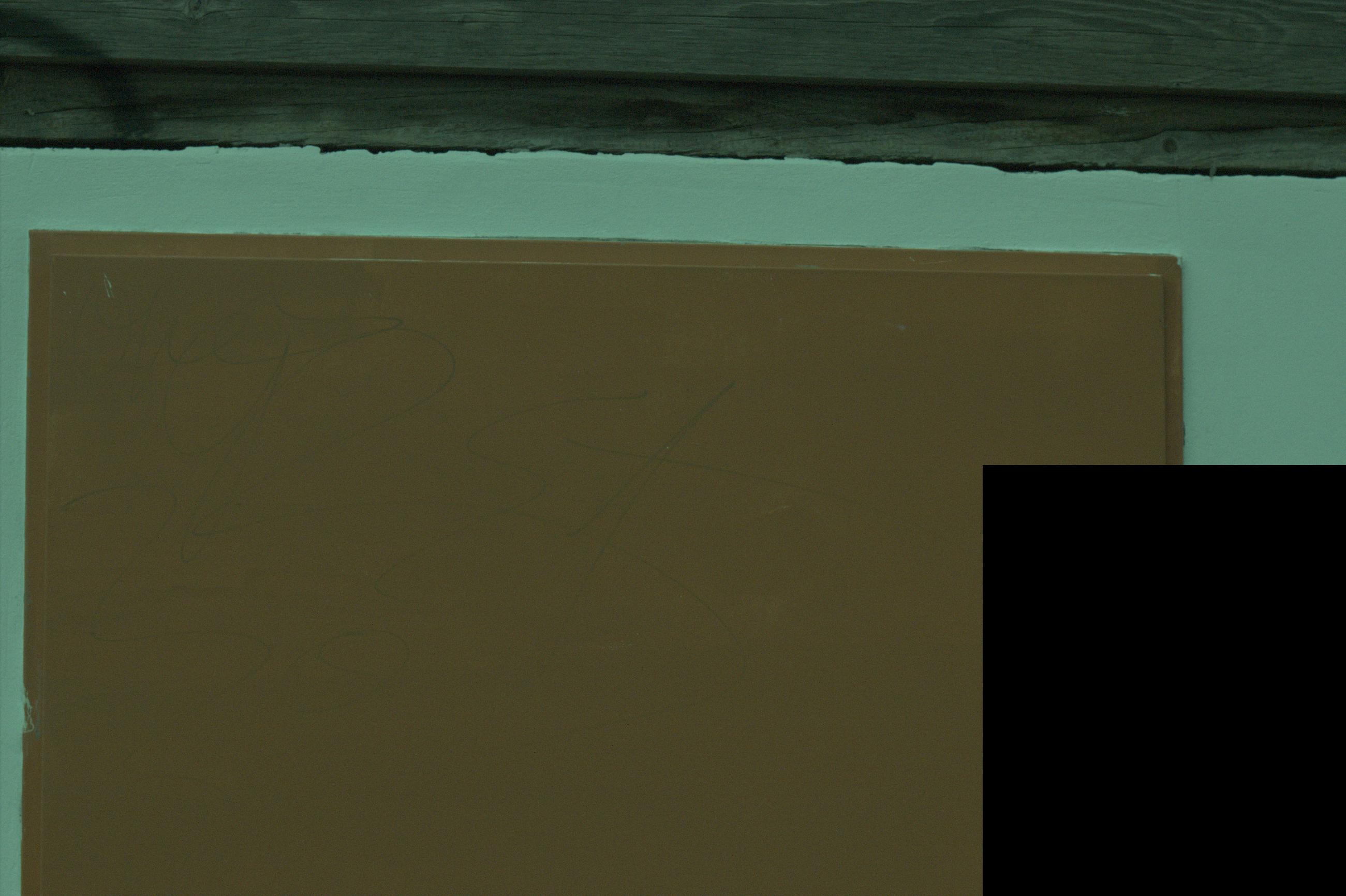}}}
&{{\includegraphics[width=2cm,height=1.4cm]{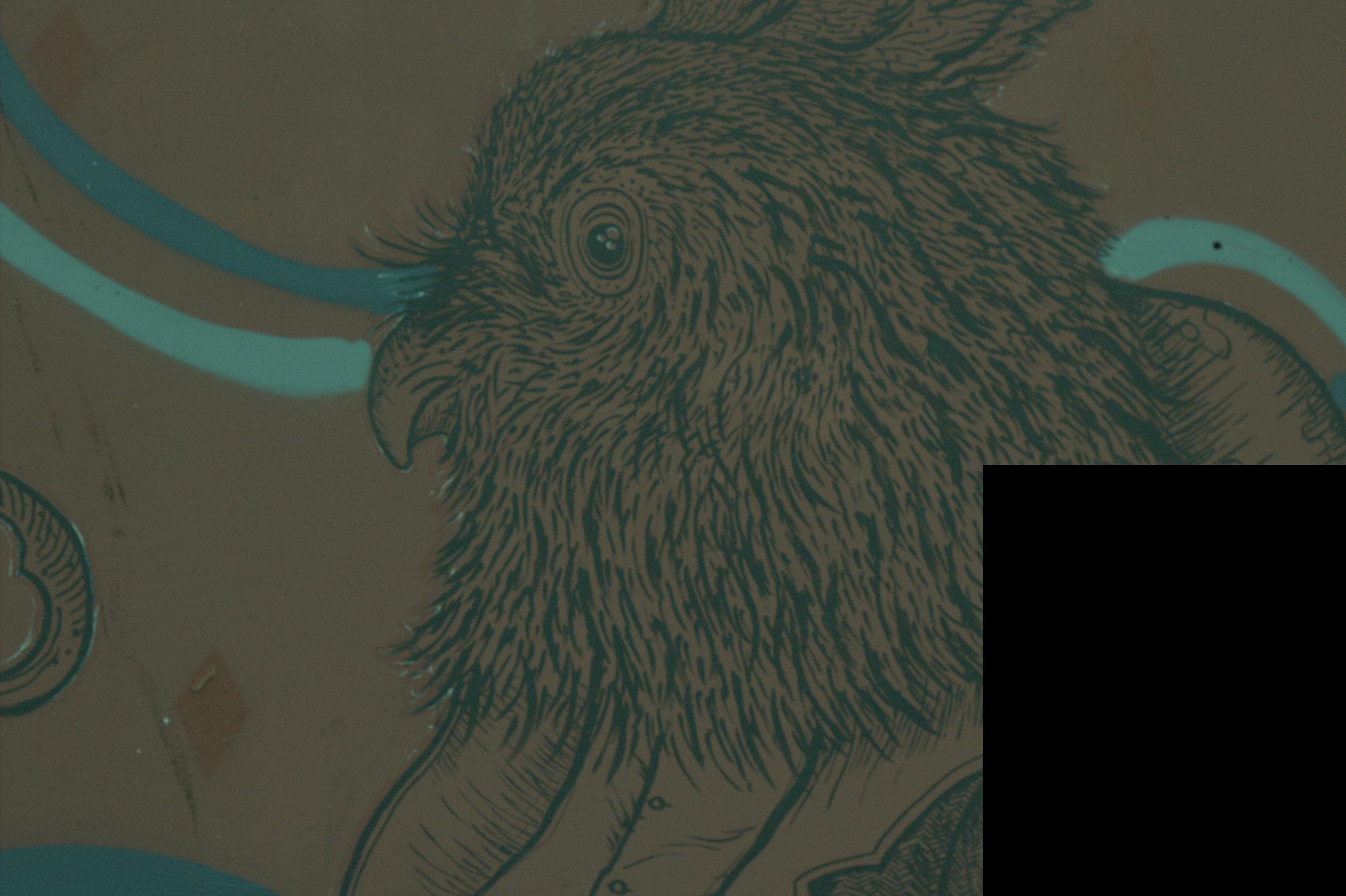}}}
& {{\includegraphics[width=2cm,height=1.4cm]{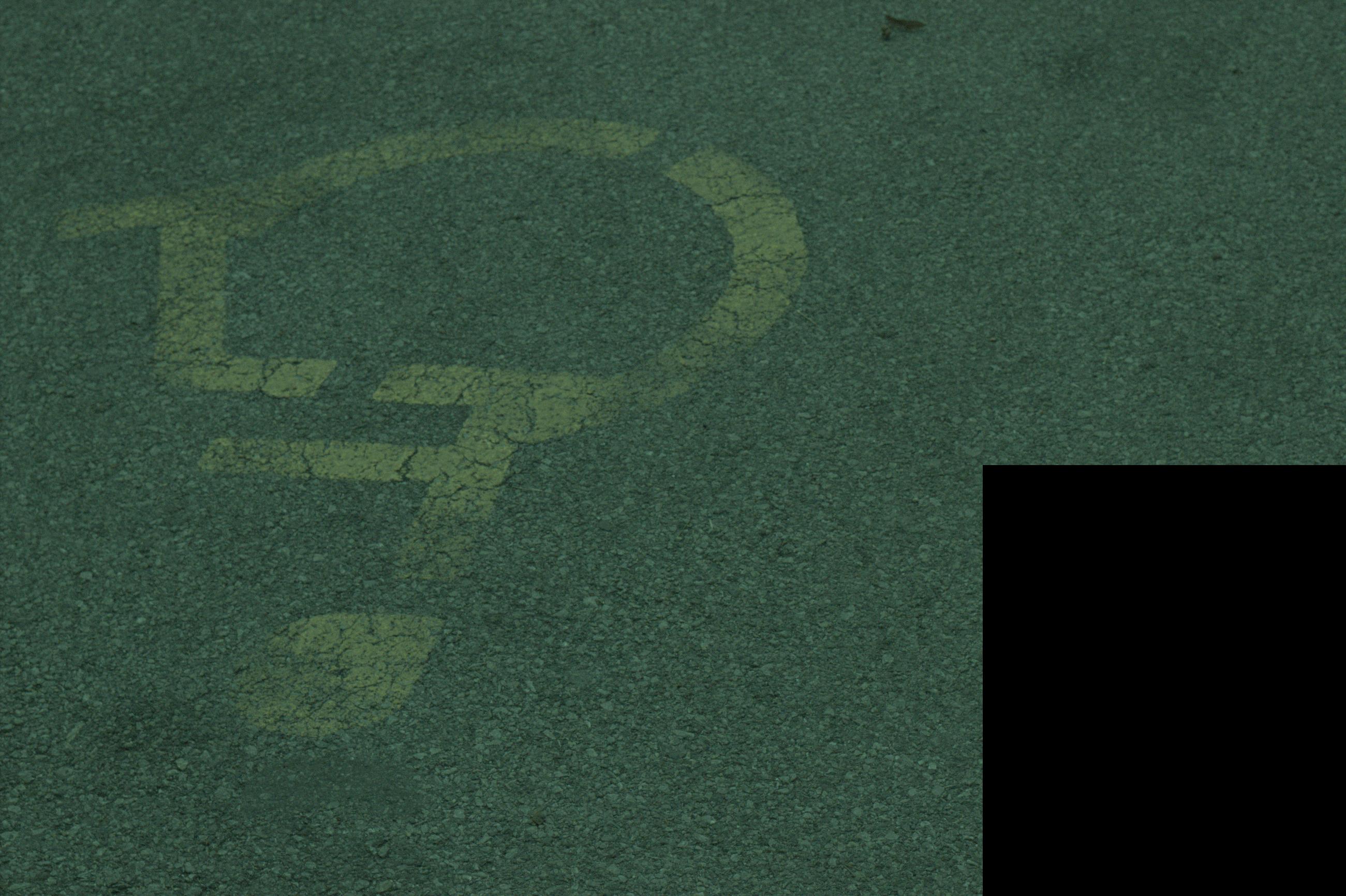}}}

\\

\vspace{-3mm}
\raisebox{2\height}{\rotatebox{0}{{  }}}
& \raisebox{2\height}{\textit{id:1}}
& \raisebox{2\height}{\textit{id:3}}
& \raisebox{2\height}{\textit{id:33}}
& \raisebox{2\height}{\textit{id:41}}
\\

\hspace{-15pt}
\vspace{-0.75mm}
\raisebox{2\height}{\rotatebox{0}{{  }}}
&{{\includegraphics[width=2cm,height=1.4cm]{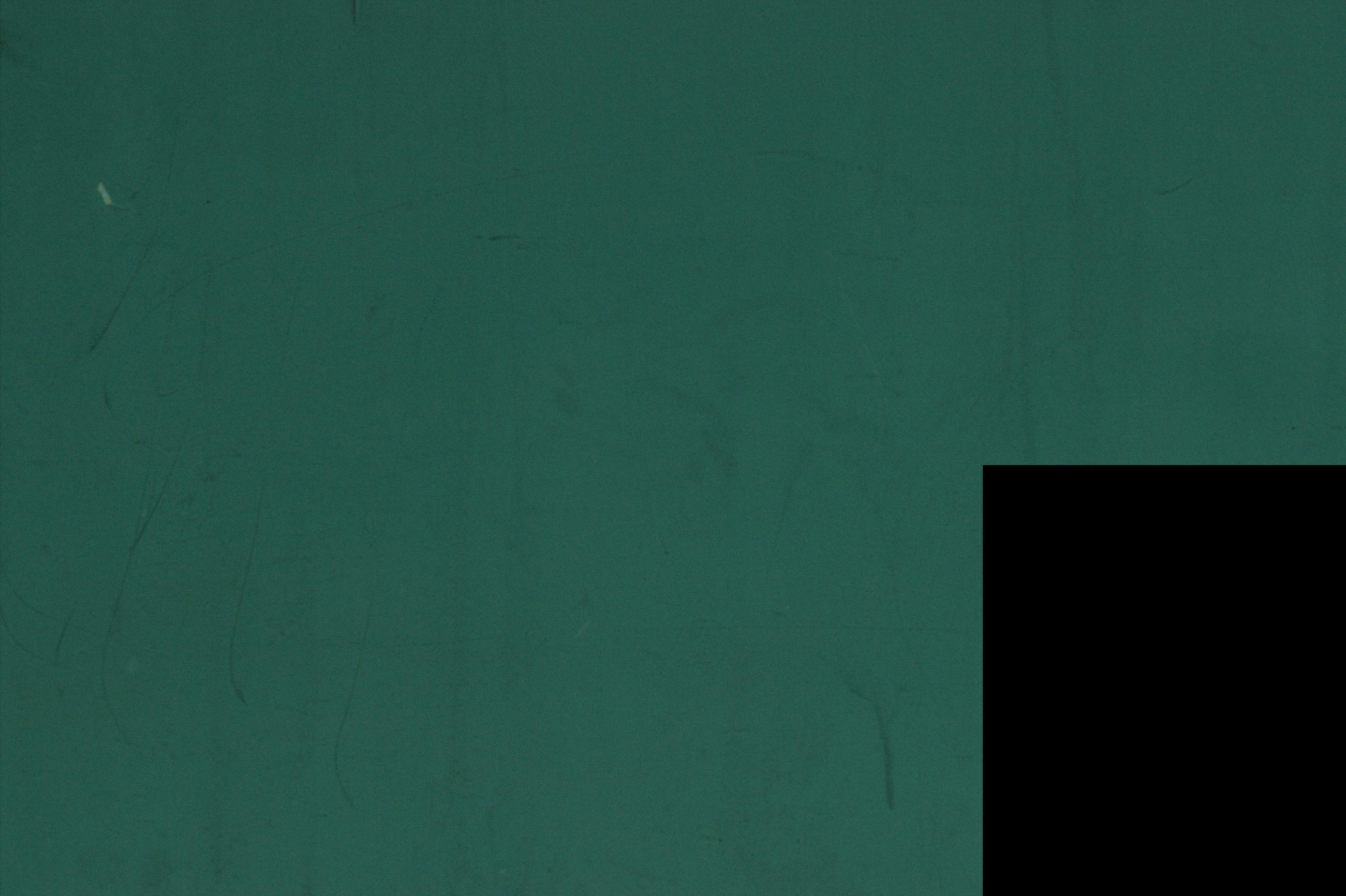}}}
& {{\includegraphics[width=2cm,height=1.4cm]{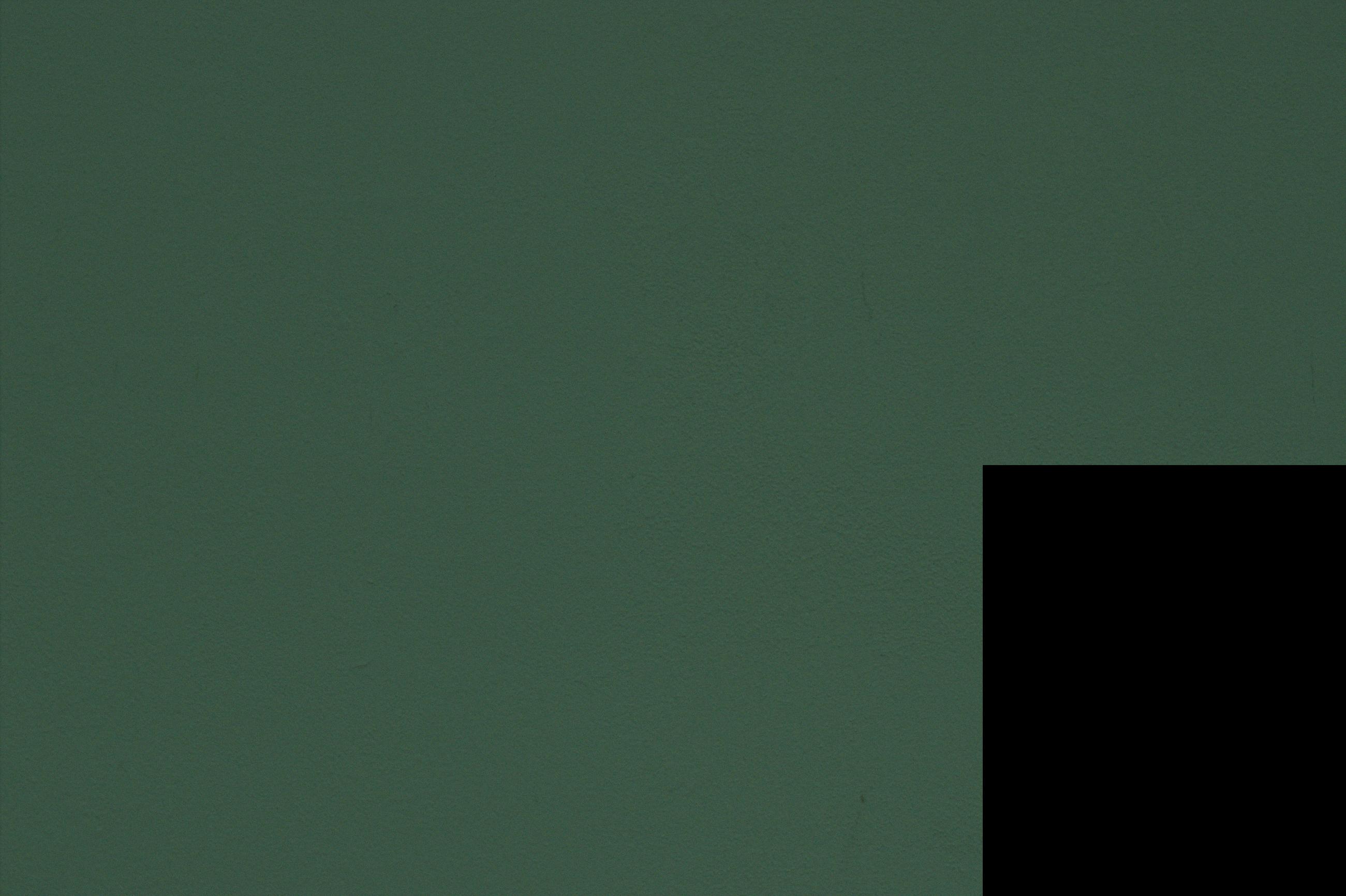}}}
&{{\includegraphics[width=2cm,height=1.4cm]{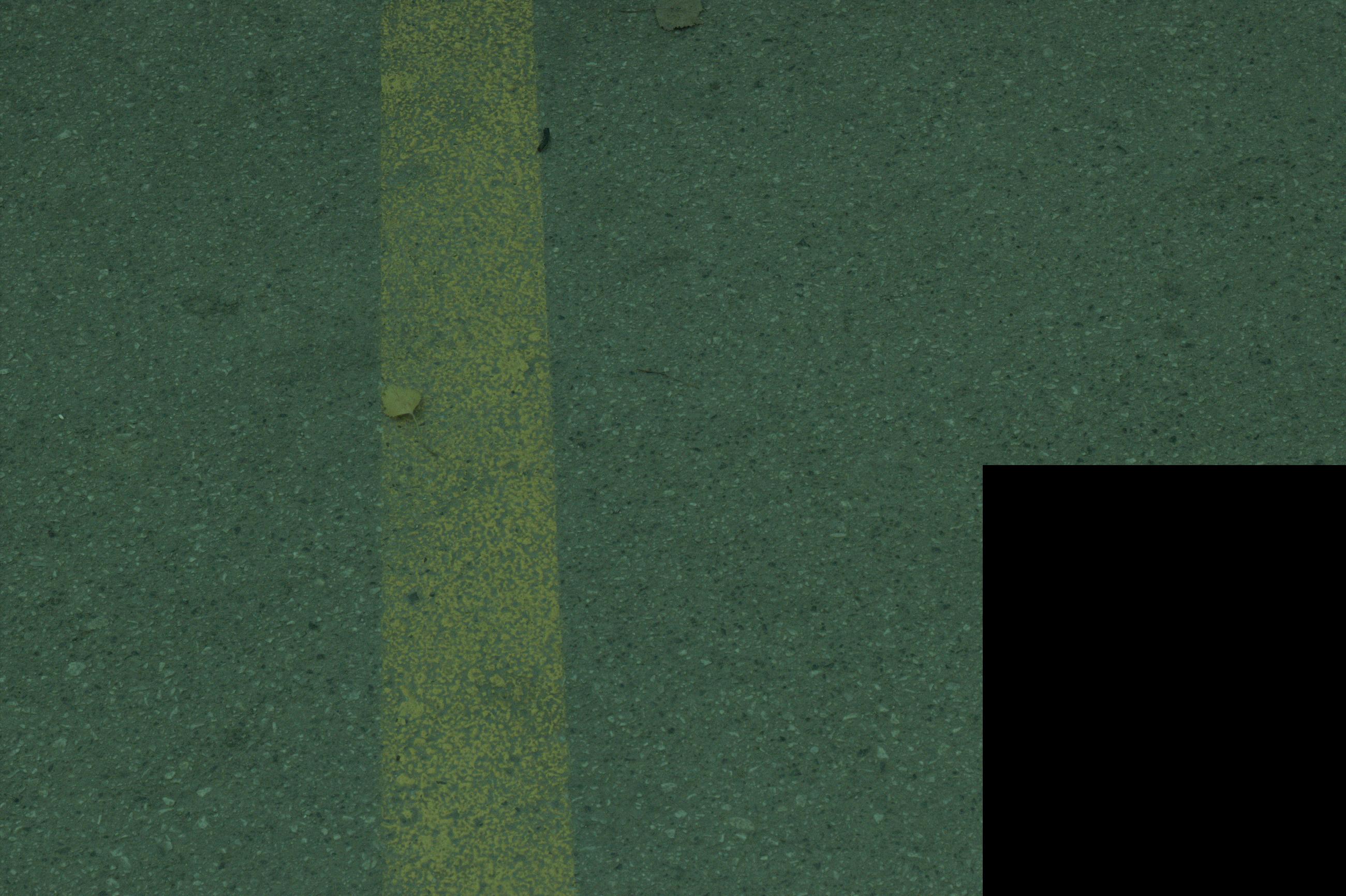}}}
& {{\includegraphics[width=2cm,height=1.4cm]{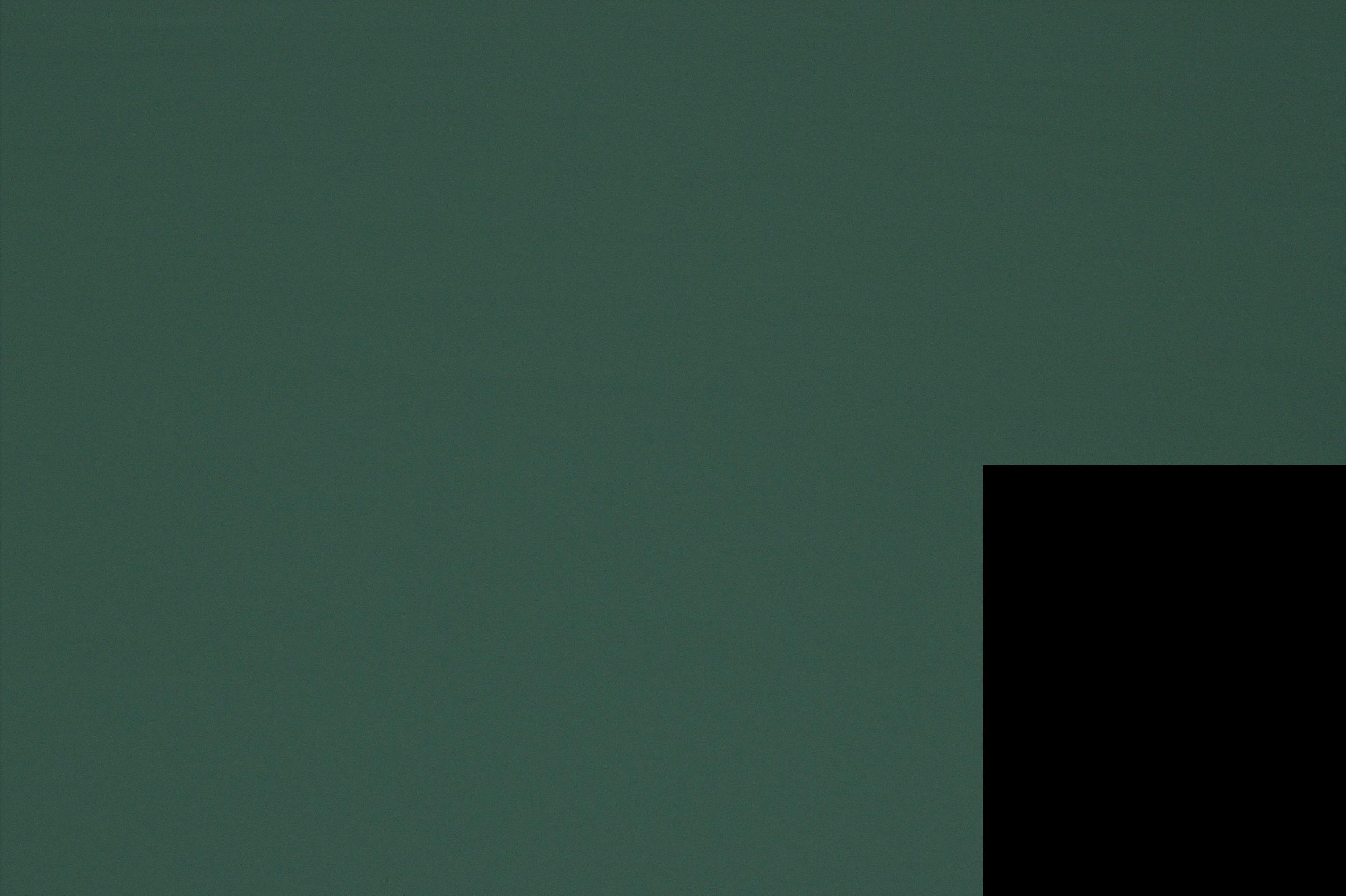}}}

\\

\vspace{-3mm}
\raisebox{2\height}{\rotatebox{0}{{  }}}
& \raisebox{2\height}{\textit{id:55}}
& \raisebox{2\height}{\textit{id:61}}
& \raisebox{2\height}{\textit{id:66}}
& \raisebox{2\height}{\textit{id:195}}

\end{tabular}

\end{center}
\vspace{-3mm}
\caption{
Hard samples from the undisclosed testing set, which can fail statistical methods like GI~\cite{yanlin2019arxiv}, while learning-based methods (FFCC~\cite{barron2017fourier} and the ranked-1st CNN method) can perform much better. Images are gamma-corrected (gamma 2.2) for better visual effects.
}
\label{figure:hard}
\vspace{-3mm}
\end{figure}

For GI-based method, it obtains a competitive median angular error, with no learning. It inherits the simplicity of the classical gray world method, but also the same drawback -- when the scene is dominated by a specific color or the image is a local patch, its result is seriously biased. This is shown by its mean error, as several ``outlier'' images can bring very high angular errors, increasing the mean error rapidly. This is consistent with the observations in \cite{yanlin2019arxiv}.
Figure~\ref{figure:hard} illustrates several hard images in the testing set for GI. On Cube+ training set which contains much less ``outlier'' images, GI obtains more accurate and robust results (Table~\ref{tab:leaderboard}, bottom). 

We also show the result of two methods tested on Gehler-Shi Dataset (Table~\ref{tab:gehlershi}) and NUS 8-camera Dataset (Table~\ref{tab:nus}).  Comparing the three tables, we find that the Cube+ dataset is the most challenging one.  It is partially due to a large portion of local-patch images (Figure~\ref{figure:hard}) in the Cube+ dataset. 

\section{Conclusion}
In this paper we test two non-deep-learning methods on the new challenging Cube+ dataset. The FFCC method needs training and can exploit the exif information (if provided), while the Grayness Index is a simple statistical methods but sensitive to extremely hard images. Both methods find their position in the leaderboard, reasonably.  Our future plan is to make Grayness Index include more dichromatic cues to deal with very local images, while still keeping it simple.

\setlength{\tabcolsep}{1pt}
\renewcommand{\arraystretch}{1}

{\small
\bibliographystyle{ieee}
\bibliography{cvpr2019}
}

\end{document}